# Fire Detection From Image and Video Using YOLOv5

Arafat Islam
*Computer Science*
*American International University- Bangladesh*
*19-39377-1@student.aiub.edu*

Md. Imtiaz Habib
*Computer Science*
*American International University- Bangladesh*
*19-39389-1@student.aiub.edu*

*Abstract*— **For the detection of fire-like targets in indoor, outdoor and forest fire images, as well as fire detection under different natural lights, an improved YOLOv5 fire detection deep learning algorithm is proposed. The YOLOv5 detection model expands the feature extraction network from three dimensions, which enhances feature propagation of fire small targets identification, improves network performance, and reduces model parameters. Furthermore, through the promotion of the feature pyramid, the top-performing prediction box is obtained. Fire-YOLOv5 attains excellent results compared to state-of-the-art object detection networks, notably in the detection of small targets of fire and smoke with mAP 90.5% and f1 score 88%. Overall, the Fire-YOLOv5 detection model can effectively deal with the inspection of small fire targets, as well as fire-like and smoke-like objects with F1 score 0.88. When the input image size is 416 × 416 resolution, the average detection time is 0.12 s per frame, which can provide real-time forest fire detection. Moreover, the algorithm proposed in this paper can also be applied to small target detection under other complicated situations. The proposed system shows an improved approach in all fire detection metrics such as precision, recall, and mean average precision.**

## 1. INTRODUCTION

Humans require fire because it provides them with energy. But at the same time, unexpected fire causes many accidents. According to the National Fire Protection Association (NFPA), fire departments respond to over 350,000 home structure fires nationwide, causing almost $7 billion in direct damage [1]. To prevent such fires, it is crucial to detect fires rapidly. Traditionally, fires have been detected using sensors, and putting sensors in every place where fire may occur is expensive as it does not cover ample space and maintenance is time-consuming. Due to rapid developments in digital camera technology and video processing techniques, there is a major trend to replace conventional fire detection methods with computer vision-based systems. With the development of machine learning, deep learning techniques have been widely used in detection [2]. The authors of [3] propose an early fire detection framework using fine-tuned Convolutional Neural Networks (CNN), but the model has a high computational cost. Fire Detection Using Image Processing is proposed in [4,5]. A forest smoke detection algorithm based on YOLO-V3 and YOLO-V4 is proposed in [6].

We propose a fire detection system based on a novel convolutional neural network (CNN) to address this need, using the YOLOv5 model [7]. The proposed system combines deep learning models and modern computer vision.

## 2. DESCRIPTION OF DATA

### 2.1 DATASET ACQUISITION

The level of precision of the deep learning model primarily depended on the dataset used in the training and validation procedures. We gathered fire images from various open access sources such as GitHub [8] and Roboflow [9], finding images depicting a range of different conditions (shape, color, size, indoor and outdoor environment). Our fire image dataset consisted of 2462 fire images, as shown in Table 1.

Table-1: Image sources for fire detection dataset.

| Dataset | Image Size | Number of images |
|---|---|---|
| Github | 300 x 168, 299 x 169<br>200 x 253, 259 x 195<br>251 x 201, 264 x 191 | 502 |
| Roboflow | 640 x 640 | 1960 |
| Total Images | | 2462 |

### 2.2 DATA PREPROCESSING

The dataset was store in roboflow. We removed images that were not annotated. And combine them into a dataset. The images were preprocessed using roboflow framework [9] We did not apply any augmentation. They were resized and stretched to 416 x 416. Finally, divide them for training (50%) and validation (50%) as shown in Table 2.

Table-2: Distribution of training and validation images in fire dataset.

| Dataset Name | Size | Number of images | Training Set | Validation Set |
|---|---|---|---|---|
| Fire-Dataset | 416 x 416 | 2427 | 1214 | 1213 |

Then we export dataset as YOLO v5 Pytorch format and it generates a API link which we used in our model directly.

## 3. MODEL

### 3.3 DESIGN OF THE PROPOSED SYSTEM

This study intends to improve fire detection from other single-stage object detector images and videos based on the YOLOv5 architecture. We have designed our system from data collection to model building. Figure 1 depicts the general architecture of the proposed system and gives a more detailed explanation. After data collection, we preprocessed data then they were split into training (50%) and validation (50%) set. Then these images were passed into the YOLOv5 model. The model was trained and validate at the same time. We have used two models YOLOv5s and YOLOv5x.

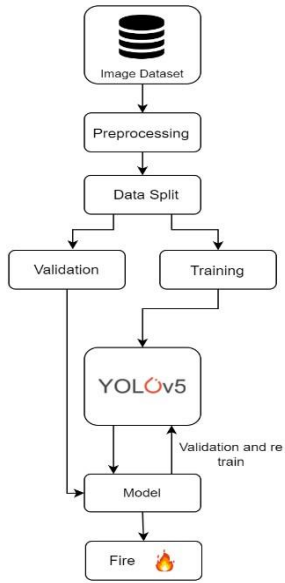

Figure-1: Overall design of the proposed system.

### 3.2 YOLOV5S NETWORK ARCHITECTURE

YOLOv5 network [10,11] is the latest product of the YOLO architecture series. The detection accuracy of this network model is high, and the inference speed is fast, with the fastest detection speed being up of 140 frames per second. On the other hand, the size of the weight file of YOLOv5 target detection network model is small, which is nearly 90% smaller than YOLOv4.

The YOLOv5 architecture contains five architectures, specifically named YOLOv5n [11], YOLOv5s [11], YOLOv5m [11], YOLOv5l [11] and YOLOv5x [11], respectively. The main difference among them is that the amount of feature extraction modules and convolution kernel in the specific location of the network is different. The accuracy, efficiency and size of the recognition model were considered comprehensively in the study, and the improved design of the fire targets recognition network was carried out based on the YOLOv5n and YOLOv5x architecture (Figure-2).

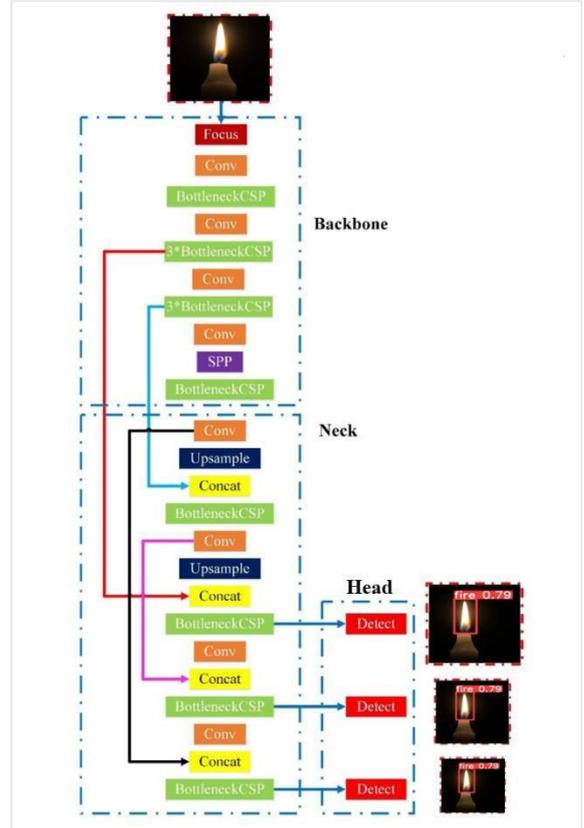

Figure-2: Architecture of the YOLOv5s network.

### 3.3 ARCHITECTURE DESIGN ANALYSIS

As YOLO v5 is a single-stage object detector, it has three important parts like any other single-stage object detector.

1. Model Backbone
2. Model Neck
3. Model Head

### 3.3.1 MODEL BACKBONE

Model Backbone is mainly used to extract important features from the given input image. In YOLO v5 the CSP-Cross Stage Partial Networks [12] are used as a backbone to extract rich in informative features from an input image.

### 3.3.2 MODEL NECK

Model Neck is mainly used to generate feature pyramids. Feature pyramids help models to generalized well on object scaling. It helps to identify the same object with different sizes and scales. The first layer of the backbone network is the focus module (Figure 3), which is designed to reduce the calculation of the model and accelerate the training speed. Its functions are as follows: Firstly, the input 3 channel image (the default input image size of YOLOv5s [11] architecture is $3 \times 414 \times 414$) was segmented into four slices with the size of $3 \times 207 \times 207$ per slice, using a slicing operation. Secondly, concat operation was utilized to connect the four sections in depth, with the size of output feature map being $12 \times 207 \times 207$, and then through the convolutional layer composed of 32 convolution kernels, the output feature map with a size of $32 \times 207 \times 207$ was generated. Finally, through the BN layer (batch normalization) and the Leaky ReLU activation functions, the results were output into the next layer.

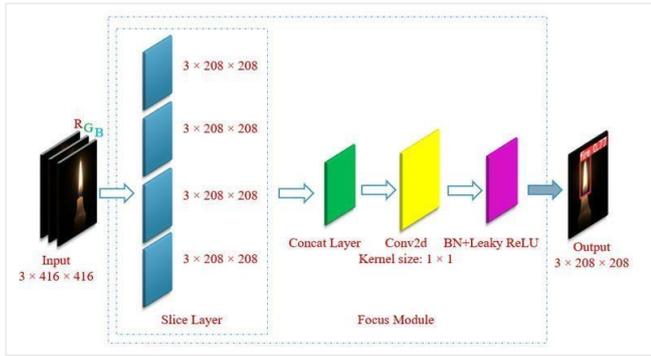

Figure 3. Structure of Focus module.

### 3.3.3 MODEL HEAD

The model Head is mainly used to perform the final detection part. It applied anchor boxes on features and generates final output vectors with class probabilities, objectness scores, and bounding boxes and sigmoid activation function is used in this detection layer.

### 3.4 ACTIVATION AND OPTIMIZATION FUNCTION

In YOLO v5 the Leaky ReLU activation function is used in middle/hidden layers and the sigmoid activation function is used in the final detection layer. Leaky-ReLU activation looks as follows:

$$R(x) = \begin{cases} x, & if\ x \geq 0 \\ \alpha x, & otherwise \end{cases}$$

The mathematical definition of the sigmoid function is as follows:

$$f(x) = \frac{1}{1 + e^{-x}}$$

In YOLO v5, the default optimization function for training is SGD.

The mathematical definition of the Stochastic Gradient Descent function is as follows:

$$\omega_{t+1} = \omega_t - \alpha \frac{\partial L}{\partial \omega_t}$$

Where, the vanilla gradient descent updates the current weight $w_t$ using the current gradient $\partial L/\partial w_t$ multiplied by some factor called the learning rate, $\alpha$.

### 3.5 COST FUNCTION OR LOSS FUNCTION

In the YOLO family, there is a compound loss is calculated based on objectness score, class probability score, and bounding box regression score.

Ultralytics [11] have used Binary Cross-Entropy with Logits
Loss function from PyTorch for loss calculation of class probability and object score.

This loss combines a *Sigmoid* layer and the *BCELoss* in one single class.

The unreduced (i.e. with reduction set to 'none') loss can be described as:

$(x,y)=L=\{l_1,...,l_N\}^T$,
$l_n = -w_n[v_n \cdot \log\sigma(x_n) + (1-v_n) \cdot \log(1-\sigma(x_n))]$

where $N$ is the batch size. If reduction is not 'none' (default 'mean'), then

$$\ell(x, y) = \begin{cases} \text{mean}(L), & \text{if reduction} = \text{'mean'}; \\ \text{sum}(L), & \text{if reduction} = \text{'sum'}. \end{cases}$$

It's possible to trade off recall and precision by adding weights to positive examples. In the case of multi-label classification, the loss can be described as:
$\ell c(x,y) = L_c = \{l_1,c,...,l_N,c\}^T$
$l_{n,c} = -w_{n,c}[p_c y_{n,c} \cdot \log\sigma(x_{n,c}) + (1-y_{n,c}) \cdot \log(1-\sigma(x_{n,c}))]$,

Where $c$ is the class number ($c>1$ for multi-label binary classification, $c=1$ for single-label binary classification), $n$ is the number of the sample in the batch and $pc$ is the weight of the positive answer for the class $c$. $pc>1$ increases the recall, $pc<1$ increases the precision.

## 4. RESULT ANALYSIS

### 4.1 PERFORMANCE INDICATORS

In the study, objective evaluation indicators such as Precision, Recall, mAP (mean average precision) and F1 score were used to evaluate the performance of the trained fire targets recognition model. The calculation equations are as follows:

$$Precision = \frac{TP}{TP + FP}$$

$$Recall = \frac{TP}{TP + FN}$$

The precision–recall curve (P–R curve) is obtained by taking the precision ratio as the vertical axis and the recall rate as the horizontal axis. The F1 score is also

used to evaluate the performance of the model. The definition of F1 score is as follows:

$$F1 = \frac{2 \times Precision \times Recall}{Precision + Recall}$$

The mAP is calculated by finding Average Precision (AP) for each class and then average over a number of classes.

$$mAP = \frac{1}{N}\sum_{i=1}^{n} AP_i$$

Where, *TP* means that the number of correctly identified fire targets; *FP* means that the number of misidentified backgrounds as fire targets; *FN* represents the number of unidentified fire targets; *C* represents the number of target categories; *N* represents the number of IOU thresholds, *K* is the IOU threshold, *P(k)* is the precision, and *R(k)* is the recall.

### 4.2 EXPERIMENTAL ANALYSIS

We measured the performance of the proposed approach by experimenting with YOLOv5x on the original fire dataset (2427 images) and compared the final precisions (Table 3). The Fire-YOLO detection model received 416 × 416pixel image as input, since the GPU performance limitations, the batch size is set to 64 for n and 32 for x, each model train 200 epochs, the initial learning rate is 0.001.

| Model | No. of Images | Size of Image | Batch Size | Epoch | Layers | Training Time (h) | Parameters |
|---|---|---|---|---|---|---|---|
| YOLOv5n | 2427 | 416 x 416 | 64 | 200 | 213 | 0.936 | 1760518 |
| YOLOv5x | | | 32 | | 444 | 3.23 | 86173414 |

Table-3: Model performance on datasets (2427 images)

### 4.3 ALGORITHM COMPARISON ANALYSIS

In order to verify the performance of the model proposed in this paper, images of fire are used as the training set. The proposed model is compared with YOLOv5n. The accuracy, recall, F1 score, Model size and mAP values are shown in Table 4.

Table-4: Accuracy performance of different model. shown in Figure 8 and 9.

| Metrics | YOLOv5n | YOLOv5x |
|---|---|---|
| Precision | 0.96 | 0.97 |
| Recall | 0.94 | 0.94 |
| F1 | 0.81 | 0.88 |
| mAP | 0.83 | 0.90 |
| Model Size | 3.8 MB | 173 MB |

The comparison of F1 curve of YOLOv5n & YOLOv5x are

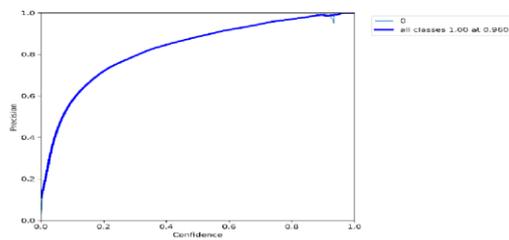

Figure-4: Precision vs Confidence (YOLOv5n)

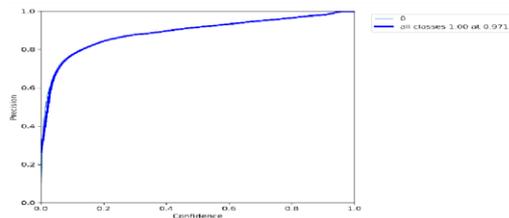

Figure-5: Precision vs Confidence (YOLOv5x)

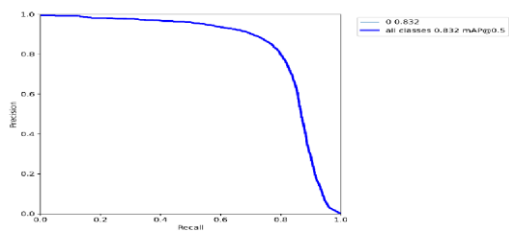

Figure-6: Precision vs Recall (YOLOv5n)

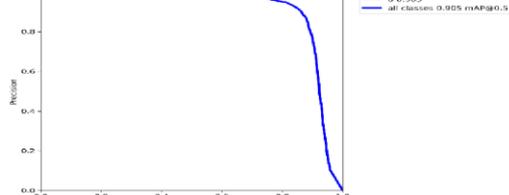

Figure-7: Precision vs Recall (YOLOv5x)

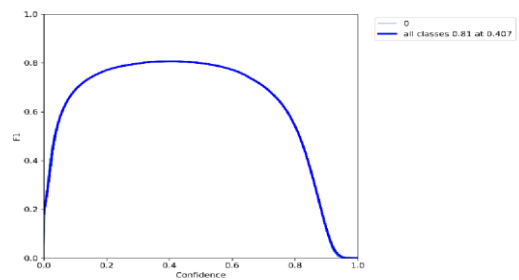

Figure-8: F1 curve of YOLOv5n

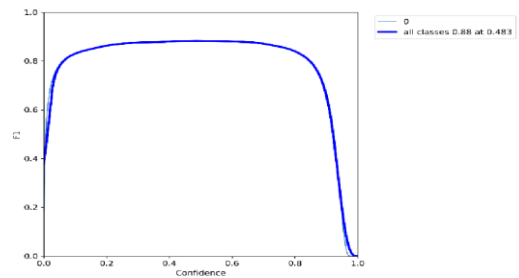

Figure-9: F1 curve of YOLOv5x

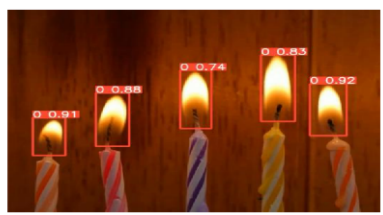

## 5. DISCUSSION

The Fire detection YOLOv5 model proposed in this paper has achieved gratifying results in small fire-like and fire detection under different lightness. It can not only provide real-time detection but also has good robustness in practical applications. Nevertheless, we found that the detection algorithm still has the problem of low detection accuracy and challenging detection of semi-occluded targets in our tests. This phenomenon may be due to the variability of fire and the complexity of fire spread in the detection of flames in the actual environment, which creates a dilemma in fire inspection. It is worth mentioning that this is also an urgent problem [13] to be solved by the current target detection model. What is encouraging is that these difficulties are not insurmountable. In fact, the corresponding deep learning training techniques can be flexibly selected in the detection algorithm, for instance, translation-invariant transformation [14], random geometric transformation [15], and random color dithering [16] on the training image in the training of the model, which shows promising results. In future work, the training process of the Fire-YOLOv5 model will be optimized, and the focus will be on the preprocessing of images. It is hoped that the generalization ability of Fire-YOLOv5 can be further improved through transfer learning.

## 6. CONCLUSIONS

In this research, a fire detection system was developed using deep CNN models YOLOv5 object detector. The proposed fire detection system was trained using two open source fire image dataset that contained different fire scenes and ran on Google Colab. It detected fire from images and videos. We created a fire detection dataset that included 2427 fire images for model training and validation. During the experiments, we evaluated the qualitative and quantitative performance of the proposed system by comparing it with other well-known one-stage object detectors. The experimental results and evaluation proved that the YOLOv5x model was robust and performed better than YOLOv5n on our fire detection dataset, with 90.5 and 83.2% mAP respectively. The proposed fire detection method is effective and can be used in various applications, allowing researcher to detect fires at an early stage. Furthermore, object mapping method can be applied to understand the system whether the fire is intentional or unintentional.

## REFERENCES

[1] Ahrens, M. and Maheshwari, R., 2021. *Home Structure Fires report | NFPA*. [online] Nfpa.org. Available at:<https://www.nfpa.org/News-and-Research/Data-researchand-tools/Building-and-Life-Safety/Home-Structure-Fires> [Accessed 12 August 2022].

[2] Zhao, C.; Feng, Y.; Liu, R.; Zheng, W. Application of Lightweight Convolution Neural Network in Cancer Diagnosis. In Proceedings of the 2020 Conference on Artificial Intelligence and Healthcare, Taiyuan, China, 23–25 October 2020; pp. 249–253.

[3] Muhammad, K.; Ahmad, J.; Baik, S.W. Early fire detection using convolutional neural networks during surveillance for effective disaster management. Neurocomputing 2018, 288, 30–42. [CrossRef]

[4] J. Seebamrungsat, S. Praising and P. Riyamongkol, "Fire detection in the buildings using image processing," 2014 Third ICT International Student Project Conference (ICTISPC), 2014, pp. 95-98, doi: 10.1109/ICT-ISPC.2014.6923226.

[5] Celik, T., 2010. Fast and Efficient Method for Fire Detection Using Image Processing. *ETRI Journal*, 32(6), pp.881-890.

[6] Jindal, P.; Gupta, H.; Pachauri, N.; Sharma, V.; Verma, O.P. Real-Time Wildfire Detection via Image-Based Deep Learning Algorithm. In Soft Computing: Theories and Applications. Advances in Intelligent Systems and Computing; Sharma, T.K., Ahn, C.W., Verma, O.P., Panigrahi, B.K., Eds.; Springer: Singapore, 2021; Volume 1381. [CrossRef]

[7] Jocher, G., Chaurasia, A., Stoken, A., Borovec, J., Kwon, Y., Fang, J., Michael, K., V, A., Montes, D., Nadar, J., Skalski, P., Wang, Z., Hogan, A., Fati, C., Mammana, L., Patel, D., Yiwei, D., You, F., Hajek, J., Diaconu, L. and Minh, M., 2022. *ultralytics/yolov5: v6.1 - TensorRT, TensorFlow Edge TPU and OpenVINO Export and Inference*. [online] Zenodo. Available at: https://zenodo.org/record/6222936#.YvYVXnZBy5e [Accessed 8 July 2022].

[8] MOSES, O., 2019. *GitHub - OlafenwaMoses/FireNET: A deep learning model for detecting fire in video and camera streams*. [online] GitHub. Available at: <https://github.com/OlafenwaMoses/FireNET/> [Accessed 1 August 2022].

[9] App.roboflow.com. 2022. [online] Available at: <https://app.roboflow.com/aiub-j9lom/collected-firedataset/1> [Accessed 1 August 2022].

[10] Liu, Y.; Lu, B.; Peng, J.; Zhang, Z. Research on the use of YOLOv5 object detection algorithm in mask wearing recognition. *World Sci. Res. J.* 2020, 6, 276–284. [Google Scholar]

[11] ultralytics. yolov5. Available online:
 https://github.com/ultralytics/yolov5 (accessed on 18 July 2022).

[12] C. -Y. Wang, H. -Y. Mark Liao, Y. -H. Wu, P. -Y. Chen, J. -W. Hsieh and I. -H. Yeh, "CSPNet: A New Backbone that can Enhance Learning Capability of CNN," 2020 IEEE/CVF Conference on Computer Vision and Pattern Recognition Workshops (CVPRW), 2020, pp. 1571-1580, doi: 10.1109/CVPRW50498.2020.00203.

[13] K. Muhammad, J. Ahmad, I. Mehmood, S. Rho and S. W. Baik, "Convolutional Neural Networks Based Fire Detection in Surveillance Videos," in IEEE Access, vol. 6, pp. 18174-18183, 2018, doi: 10.1109/ACCESS.2018.2812835.

[14] Luo, D.; Wang, D.; Guo, H.; Zhao, X.; Gong, M.; Ye, L. Detection method of tubular target leakage based on deep learning. In Proceedings of the Seventh Symposium on Novel Photoelectronic Detection Technology and Application, Kunming, China, 5–7 November 2020; Volume 11763, p. 1176384

[15] Mumuni, A., Mumuni, F. CNN Architectures for Geometric Transformation-Invariant Feature Representation in Computer Vision: A Review. *SN COMPUT. SCI.* 2, 340 (2021).

[16] Kayhan, O.S.; Gemert, J.C. On translation invariance in cnns: Convolutional layers can exploit absolute spatial location. In Proceedings of the IEEE/CVF Conference on Computer Vision and Pattern Recognition, Seattle, WA, USA, 13–19 June 2020 pp. 14274–14285